\documentclass[letterpaper]{article} \usepackage[preprint]{aaai2027}  \usepackage[hyphens]{url}  \usepackage{graphicx}    \usepackage{natbib}  \usepackage{caption} 

\usepackage{algorithm}
\usepackage[noEnd, indLines]{algpseudocodex}

\usepackage{xspace}

\pgfset{algpxIndentLine/.style={draw=black, thin}}

\usepackage{xcolor}

\newcommand{\suregen}{\texttt{SureGen}\xspace}
\newcommand{\suregenfw}{\texttt{SureGen-fw}\xspace}
\newcommand{\suregenbt}{\texttt{SureGen-bt}\xspace}
\newcommand{\rs}{\texttt{RS}\xspace}
\newcommand{\srs}{\texttt{SRS}\xspace}
\newcommand{\rsc}{\texttt{RS\_C}\xspace}
\newcommand{\srsc}{\texttt{SRS\_C}\xspace}
\newcommand{\srnds}{\texttt{S\_RND\_S}\xspace}
\newcommand{\pconflict}{\textsc{P-Conflict}\xspace}
\newcommand{\punsat}{\textsc{P-Unsat}\xspace}

\makeatletter
\algnewcommand\algorithmicinvariant{\textbf{Invariant:}}
\algnewcommand\Invariant{  \algpx@endCodeCommand  \ifnumcomp{0}{<}{\FSSize{algpx@startNewCodeBoxQueue}}{\setbool{algpx@setNorth}{true}}{}  \algpx@drawInItem{\algorithmicinvariant}}
\makeatother

\usepackage{booktabs}

\usepackage{amsmath}
\usepackage{amssymb}
\usepackage{mathtools}
\usepackage{amsthm}
\usepackage{multirow}
\usepackage{array}
\usepackage{makecell}
\usepackage{tikz}
\usetikzlibrary{positioning, arrows.meta, shapes, backgrounds, shapes.geometric, calc}
\usepackage{xcolor}
\usepackage{tcolorbox}
\tcbuselibrary{listings}
\lstdefinestyle{pythonexample}{
    language=Python,
    basicstyle=\small\ttfamily,
    keywordstyle=\color[HTML]{008000}\bfseries,
    stringstyle=\color[HTML]{BA2121},
    commentstyle=\color[HTML]{3D7B7B}\itshape,
    emph={LaptopConfig},
    emphstyle=\color[HTML]{0000FF}\bfseries,
    columns=fullflexible,
    keepspaces=true,
    showstringspaces=false,
    upquote=true,
    tabsize=4
}
\usepackage{enumitem}
\usepackage{threeparttable}
\usepackage{subcaption}

\theoremstyle{plain}
\newtheorem{theorem}{Theorem}

\theoremstyle{definition}
\newtheorem{definition}[theorem]{Definition}

\theoremstyle{remark}

\title{Specification-Guided Generation for Large Language Models}
\author{
    Alessandro Previti\thanks{Corresponding author}\textsuperscript{1},
    Konstantinos Vandikas\textsuperscript{1},
    Mario Alviano\textsuperscript{2},
    Mats Carlsson\textsuperscript{3},
    Albin Larsson Forsberg\textsuperscript{1},
    Alexandros Nikou\textsuperscript{1},
    Athanasios Karapantelakis\textsuperscript{1},
    Aneta Vulgarakis Feljan\textsuperscript{1}
}
\affiliations{
    \textsuperscript{1}Ericsson AB, Stockholm, Sweden\\
    \textsuperscript{2}DEMACS, University of Calabria, Rende, Italy\\
    \textsuperscript{3}RISE Research Institutes of Sweden, Stockholm, Sweden\\
    \{alessandro.previti, konstantinos.vandikas, albin.larsson.forsberg,\\
    alexandros.nikou, athanasios.karapantelakis, aneta.vulgarakis\}@ericsson.com\\
    mario.alviano@unical.it, mats.carlsson@ri.se
}

\begin{document}

\maketitle

\begin{abstract}
Large language models (LLMs) can generate structured outputs, but existing
constrained-decoding methods mainly enforce syntax and cannot directly enforce
satisfaction of a formal specification. We introduce a neuro-symbolic,
specification-guided decoding framework that combines LLM generation with symbolic
reasoning. At each step, the framework refines the decoding grammar by filtering out
values that cannot be extended to an output satisfying the specification. We present a
forward-only variant that guarantees specification satisfaction by construction, and a
backtracking variant that revises earlier feasible choices when their downstream
consequences leave only values that the LLM itself judges unsuitable. The framework is
model-agnostic, requires no retraining, and builds on existing grammar-constrained
decoding mechanisms. Experiments on configuration and constraint-solving benchmarks show
that the forward-only variant reliably produces specification-compliant outputs while
reducing repeated LLM inference compared with rejection sampling and incurring modest
symbolic-reasoning overhead. The backtracking variant further reduces cases in which the
model is forced to select values that it has itself judged unsuitable, thereby
substantially improving adherence to the LLM’s stated interpretation of the prompt.
\end{abstract}

\section{Introduction}

The success of Large Language Models (LLMs)~\cite{BrownMRSKDNSSAA20,
Touvron23, Gem23, Guo25} has motivated constrained-generation
methods~\cite{PoesiaP00SMG22, pmlr-v235-beurer-kellner24a, ParkWBPD24},
which restrict decoding so that only tokens preserving syntactic validity are
admissible. These methods guarantee conformity to regular expressions,
context-free grammars, or structured schemas without model retraining, making
them useful for applications such as API generation, information extraction,
code synthesis, and tool invocation.

However, existing constrained-decoding methods are limited to syntactic
structure and cannot directly enforce a formal specification expressed as
logical constraints.
Here, a formal specification is
a mathematically precise description of the properties that every admissible
output must satisfy.
As a result, generated outputs may be syntactically valid while still violating
logical relationships and dependencies among variables.
Such guarantees are increasingly
important as LLMs are used for monitoring, configuration, and decision-making in
domains including telecommunications, software systems, healthcare, and
finance~\cite{LiGLGLWHL25, 11143399, cai2025refinement, WuICLR24,
singhal2023clinical, liu2023fingpt, 0036YZXLL0DMYZ024}. 

We introduce \emph{specification-guided decoding}, which integrates a symbolic
reasoner directly into LLM generation. Before each assignment, the reasoner
computes the values that remain compatible with the specification and restricts
decoding to those values. Consequently, every completed output satisfies the
formal specification by construction.
Unlike approaches that translate natural-language requirements into formal
constraints~\cite{YeCDD23, WuICLR24, pmlr-v267-zhang25ds, PanAWW23}, we assume
that the specification is provided as input, analogously to the predefined
grammar or schema required by structured-output tools. Our approach is therefore
complementary to specification-synthesis methods, is model-agnostic, requires no
retraining, and builds on existing constrained-decoding mechanisms.

The LLM and the symbolic reasoner have complementary roles. The specification
captures hard requirements, while the prompt expresses the user's goals and
preferences within the feasible space. The reasoner guarantees validity, whereas
the LLM guides the choice among potentially many valid outputs toward one that
better reflects the prompt.
The LLM’s preferences over feasible values are context-dependent and revealed only during
autoregressive generation, so they cannot be fully formalized in advance. Moreover, the
constrained variables may form only part of a broader grammar-guided output, and their
preferred values may depend on surrounding content generated during decoding. Solving the
specification independently would therefore neither preserve these context-dependent
choices nor replace the overall generation process. Our procedure instead preserves the
model’s native generation process as much as possible: similarly to structured-output
decoding, it intervenes only to exclude infeasible choices while otherwise following the
model’s conditional distribution.

\begin{figure}[!t]
\centering
\begin{tcolorbox}[
  width=\columnwidth,
  colback=gray!5,
  colframe=black,
  boxrule=0.5pt,
  arc=2pt,
  left=6pt,
  right=6pt,
  top=6pt,
  bottom=6pt
]

\begin{lstlisting}[style=pythonexample,basicstyle=\fontsize{8}{9.5}\selectfont\ttfamily]
class LaptopConfig(BaseModel):
    CPU: Literal['i5','i7','i9']
    RAM: Literal['ram8','ram16','ram32']
    GPU: Literal['integrated','dedicated']
    HARD_DISK: Literal[256,512,1024]
    BATTERY: Literal['standard','extended']
\end{lstlisting}

\vspace{0.5em}
Constraints:
\begin{enumerate}[noitemsep, topsep=2pt, leftmargin=1.5em]\footnotesize
    \item If GPU is Dedicated, Battery must be Extended.
    \item If CPU is an i7 or i9, RAM must be at least 16GB.
    \item If RAM is 16GB or more, Battery must be Extended.
\end{enumerate}

\end{tcolorbox}
\caption{A JSON schema expressed as a Pydantic model, with additional relational constraints.}
\label{fig:jschema}
\end{figure}

Figure~\ref{fig:jschema} illustrates the distinction. Existing
schema-constrained tools~\cite{lundberg2023guidance, llamacpp2023, willard2023}
can ensure that each laptop parameter takes an allowed value, but cannot enforce
dependencies among parameters. For example,
$CPU=i7$, $RAM=ram8$, $GPU=dedicated$, $HARD\_DISK=512$, and
$BATTERY=standard$ is syntactically valid, yet violates both the CPU--RAM and
GPU--battery constraints. Our method prevents such assignments while retaining
the LLM's ability to interpret a high-level request, such as generating a gaming
laptop configuration.

The same setting arises in information extraction and structured decision-making tasks such as
product configuration, software feature selection, scheduling, routing, and package management,
where outputs must satisfy relationships among variables.

\paragraph{Main Contributions}
We make the following contributions:
\begin{itemize}
    \item We introduce a model-agnostic specification-guided decoding framework
    that integrates symbolic reasoning into LLM generation without retraining.

    \item We develop a forward-only procedure that restricts each assignment to
    feasible values, guaranteeing specification satisfaction by construction
    without post-hoc validation or trial-and-error generation.

    \item We introduce a backtracking extension that revises earlier assignments
    when their consequences conflict with the model's interpretation of the
    prompt, while preserving specification compliance.

    \item We evaluate both methods on configuration and constraint-solving
    benchmarks. Forward decoding reduces repeated LLM inference with modest
    reasoning overhead, while backtracking resolves many prompt conflicts and
    improves adherence to the model's stated preferences.
\end{itemize}

\section{Preliminaries}

In this section we present the basic notation used throughout the paper.

\paragraph{SMT}

We use \emph{Satisfiability Modulo Theories} (SMT) to express formal
specifications; see~\cite{barrett2018satisfiability} for background. Our
approach is not specific to SMT and applies to any formalism whose solver can
check satisfiability and refine variable domains.

Let $\mathcal{F}$ be a formula over variables $\mathcal{X}$, where each
$x \in \mathcal{X}$ has domain $dom(x)$. A possibly partial assignment
$\mu$ maps variables to values in their domains and satisfies $\mathcal{F}$,
written $\mu \models \mathcal{F}$, when $\mathcal{F}$ is true under $\mu$. We
write $vars(\mathcal{F})$ for the free variables of $\mathcal{F}$.

\begin{definition}\label{def:feas-domain}
The \emph{feasible domain} of $x$ under $\mathcal{F}$ is
\[
\mathit{dom}_{\mathcal{F}}(x)
=
\{\,v \in dom(x)
\mid
\exists \mu:\mu \models \mathcal{F}\land \mu(x)=v
\,\}.
\]
\end{definition}

For example, if $dom(x)=\mathbb{Z}$ and
$\mathcal{F}$ contains $x+y<4$, $0<x<5$, and $0<y<3$, then
$dom_{\mathcal{F}}(x)=\{1,2\}$.

\paragraph{Grammar-Constrained Decoding}

Let $\mathcal{V}$ be the language-model vocabulary and let $\mathcal{G}$ be a grammar defining the
valid outputs $\mathcal{L}(\mathcal{G})$. Grammar-constrained decoding~\cite{PoesiaP00SMG22,
pmlr-v235-beurer-kellner24a, ParkWBPD24} masks tokens that would prevent the generated prefix from
being extended to a string in $\mathcal{L}(\mathcal{G})$, and selects the next token from the
resulting constrained distribution. In the remainder of this paper, we assume greedy decoding and
write $CD(p,\mathcal{G})$ for greedy grammar-constrained generation from prompt $p$
under grammar $\mathcal{G}$.
Our method applies this mechanism sequentially. Before generating an assignment
for each variable, it computes the variable's feasible domain under the formal
specification and the assignments generated so far, and constructs a grammar
that accepts exactly the values in that domain.

\section{Specification-Guided Decoding}
\label{sec:specification-guided-decoding}

In this section, we present the core idea of our approach. 
Specification-guided decoding integrates symbolic reasoning into generation so that each
decoding step preserves compatibility with the formal specification. As a result, every
partial output remains extendable to a satisfying completion, avoiding post-hoc rejection.

Let $vars(\mathcal{F})=\{x_1,\ldots,x_n\}$ be the set of variables of a
satisfiable specification $\mathcal{F}$. We denote by
$\mathcal{A}_k = \langle(x_1,v_1),\ldots,(x_k,v_k)\rangle$
the ordered trail after $k$ assignments. The residual specification induced by
the trail is
\[
\mathcal{F}_k
=
\mathcal{F}
\land
\bigwedge_{(x,v)\in A_k} (x = v),
\qquad
\mathcal{F}_0=\mathcal{F}.
\]
We denote by $\omega_k$ the output prefix containing the first $k$
assignments, and by $\mathcal{L}(\mathcal{F})$ the language of complete
assignment sequences whose corresponding assignments satisfy
$\mathcal{F}$.
Our method operates at the assignment level, with each assignment generated by
the constrained-decoding procedure $CD$. We next define the token-level
greedy decoding performed internally by $CD$.

\begin{definition}[Token-level specification-guided greedy decoding]
\label{def:spec-decoding}
Given a current token prefix $\tau_{<t}$, specification-guided greedy decoding
selects
\[
\tau_t^*
=
\mathop{\mathrm{arg\,max}}_{\tau \in \mathcal{A}(\tau_{<t}^*,\mathcal{F})}
P(\tau \mid \tau_{<t}^*),
\]
where $P(\tau \mid \tau_{<t})$ is the LLM's conditional probability of token
$\tau$ given the current prefix, and
$\mathcal{A}(\tau_{<t},\mathcal{F})$ is the set of tokens whose addition
still permits a completion satisfying $\mathcal{F}$.
\end{definition}
The function $CD$ applies this rule repeatedly until it produces a
complete value for the current variable. The resulting value constitutes the
next assignment in the assignment-level procedure.
Before assigning a variable $x$, the symbolic reasoner computes its feasible
domain under the current residual specification. The feasible domain is then
translated into a grammar $\mathcal{G}_{\mathcal{F}_k,x}$ such that
$
\mathcal{L}(\mathcal{G}_{\mathcal{F}_k,x})
=
\left\{
enc(v)
\;\middle|\;
v\in dom_{\mathcal{F}_k}(x)
\right\}$,
where $enc(v)$ denotes the textual representation of $v$.
Feasible domains are represented compactly whenever possible: integer and real-valued domains are
typically stored as lists of intervals, whereas Boolean domains and small enumerated domains are
represented as explicit sets of values.
The grammar therefore acts as an interface between the symbolic reasoner and
the language model: the reasoner identifies the feasible values, while the
grammar prevents the decoder from generating any other value.

\begin{algorithm}[t]
\caption{Specification-Guided Decoding}
\label{alg:suregenfw}
\begin{algorithmic}[1]
\Require Satisfiable specification $\mathcal{F}$, prompt $p$ 
\Ensure $\omega_n \in \mathcal{L}(\mathcal{F})$
\Invariant $\mathcal{F}_k$ is satisfiable

\State $k \leftarrow 0$
\State $\mathcal{A}_0 \leftarrow \emptyset$
\State $\mathcal{U}_0 \leftarrow \langle x_n,\ldots,x_1\rangle$
\State $\omega_0 \leftarrow \epsilon$

\While{$\mathcal{U}_k \neq \emptyset$}
    \State $k \leftarrow k+1$
    \State $(\mathcal{U}_k,x_k)
        \leftarrow pop(\mathcal{U}_{k-1})$
    \State $dom_{\mathcal{F}_{k-1}}(x_k)
        \leftarrow
        feasible\_domain(x_k,\mathcal{F}_{k-1})$

    \If{$dom_{\mathcal{F}_{k-1}}(x_k)=\{v\}$}
        \State $v_k \leftarrow v$
    \Else
        \State $\mathcal{G}_{\mathcal{F}_{k-1},x_k}
            \leftarrow
            mk\_grammar
            \bigl(dom_{\mathcal{F}_{k-1}}(x_k)\bigr)$
        \State $v_k \leftarrow
            CD
            \bigl(p\cdot\omega_{k-1}, \mathcal{G}_{\mathcal{F}_{k-1},x_k}\bigr)$
    \EndIf

    \State $\mathcal{A}_k
        \leftarrow
        \mathcal{A}_{k-1}\cdot\langle(x_k,v_k)\rangle$
    \State $\omega_k
        \leftarrow
        \omega_{k-1}\cdot(x_k=v_k)$
\EndWhile

\State \textbf{return} $\omega_n$
\end{algorithmic}
\end{algorithm}

Algorithm~\ref{alg:suregenfw} processes the variables in the chosen order. At
each iteration, it computes the feasible domain of the next variable. If the
domain is a singleton, the unique value is propagated directly, avoiding an
unnecessary call to $CD$. 
We refer to assignments obtained from singleton domains as \emph{propagated implied values}.
Otherwise, the algorithm constructs a grammar
that accepts exactly the feasible values and invokes constrained decoding to
select one of them. The selected assignment is then appended to both the trail $\mathcal{A}_{k-1}$
and the generated output $\omega_{k-1}$.
The invariant follows because
$v_k \in dom_{\mathcal{F}_{k-1}}(x_k)$, and therefore
$\mathcal{F}_k
=
\mathcal{F}_{k-1}\land(x_k=v_k)$
is satisfiable at every iteration. Hence every generated partial assignment
can be extended to a model of $\mathcal{F}$, and the final output belongs to
$\mathcal{L}(\mathcal{F})$. The procedure is therefore sound and
backtrack-free. It performs at most one call to $CD$ per variable and
requires no such call when the corresponding feasible domain is a singleton.

Regarding completeness, the procedure excludes only values that cannot be
extended to a satisfying assignment and therefore does not eliminate any
solution of the current residual specification.
This does not imply that every satisfying assignment
can be generated, since the LLM may assign zero probability to some admissible choices.

\begin{table*}[!t]
\centering
\begin{threeparttable}
\begin{tabular}{|c|c|c|c|}
\hline
 & Initial Domain & Feasible Domain & Assignment \\ \hline
{\small CPU}
& [i5, i7, i9]
& [i5, i7, i9]
& i7 \\ \hline
{\small RAM}
& [ram8, ram16, ram32]
& [ram16, ram32]
& ram16 \\ \hline
{\small GPU}
& [integrated, dedicated]
& [integrated, dedicated]
& dedicated \\ \hline
{\small HARD\_DISK}
& [256, 512, 1024]
& [256, 512, 1024]
& 1024 \\ \hline
{\small BATTERY}
& [standard, extended]
& [extended]
& extended$^{*}$ \\ \hline
\end{tabular}
\caption{Execution trace for the laptop configuration example. Prompt: \textit{``Generate a configuration for a gaming laptop.''}}
\label{tab:laptop-table}

\end{threeparttable}
\end{table*}

Table~\ref{tab:laptop-table} illustrates the execution of
Algorithm~\ref{alg:suregenfw} on the laptop configuration introduced in
Figure~\ref{fig:jschema}. The model first assigns $i7$ to $CPU$. Under
this assignment, the feasible domain of RAM is reduced to
$[ram16, ram32]$, from which the model selects $ram16$. 
The GPU variable is
not further constrained, and the LLM chooses $dedicated$,
consistent with the gaming requirement in the prompt.
Similarly, $HARD\_DISK$ remain unconstrained by the previous assignments, and the model
selects $1024$. Finally, the
specification reduces the feasible domain of $BATTERY$ to the singleton
$[extended]$; consequently, this value is propagated directly without
an additional decoding call and marked *.

\subsection{Specification-Guided Decoding with Backtracking}

\begin{algorithm}[t]
\caption{Specification-Guided Decoding with Backtracking}
\label{alg:suregen-bt}
\begin{algorithmic}[1]
\Require Specification $\mathcal{F}$, Prompt $p$
\Ensure $\omega_n \in \mathcal{L}(\mathcal{F})$ or \punsat
\Invariant $\mathcal{F}_k$ is satisfiable at the beginning of each iteration

\State $k \leftarrow 0$
\State $\mathcal{A}_0 \leftarrow \emptyset$
\State $\mathcal{U}_0 \leftarrow \langle x_n,\ldots,x_1\rangle$
\State $\omega_0 \leftarrow \epsilon$

\While{$\mathcal{U}_k\neq\emptyset$}
    \State $k \leftarrow k+1$
    \State $(\mathcal{U}_k,x_k)\leftarrow pop(\mathcal{U}_{k-1})$
    \State $dom_{\mathcal{F}_{k-1}}(x_k) \leftarrow feasible\_domain(x_k, \mathcal{F}_{k-1})$
    \If{$dom\_probe\_accepts(p, dom_{\mathcal{F}_{k-1}}(x_k))$}
        \State $\mathcal{G}_{\mathcal{F}_{k-1},x_k} \leftarrow mk\_grammar(dom_{\mathcal{F}_{k-1}}(x_k))$
        \State $v_k \leftarrow CD(p\cdot\omega_{k-1}, \mathcal{G}_{\mathcal{F}_{k-1},x_k})$
        \State $\mathcal{A}_k \leftarrow \mathcal{A}_{k-1}\cdot\langle(x_k,v_k)\rangle$
        \State $\omega_k \leftarrow \omega_{k-1}\cdot(x_k=v_k)$
    \Else
        \State $\mathcal{F} \gets \mathcal{F} \land \neg\phi_{x_k}\!\left(dom_{\mathcal{F}_{k-1}}(x_k)\right)$
        \State $\mathcal{U}_{k-1} \gets \mathcal{U}_k \cdot \langle x_k\rangle$
        \State $k \gets k-1$
        \While{$\neg SAT(\mathcal{F}_k)$}
            \If{$k > 0$}
                \State $(\mathcal{A}_{k-1},(x_k,v_k)) \gets pop(\mathcal{A}_k)$
                \State $\omega_{k-1} \gets pop(\omega_k)$
                \State $\mathcal{U}_{k-1} \gets \mathcal{U}_k \cdot \langle x_k\rangle$
                \State $k \gets k-1$
            \Else
                \State \textbf{return} \punsat
            \EndIf
        \EndWhile
    \EndIf
\EndWhile

\State \textbf{return} $\omega_n$

\end{algorithmic}
\end{algorithm}

In this section, we introduce a backtracking extension for resolving cases where
specification-compliant decoding forces values that the LLM itself judges unsuitable for
the prompt.
Algorithm~\ref{alg:suregenfw} guarantees satisfaction of the formal
specification, but an early greedy assignment may unnecessarily restrict the
feasible domains of later variables and eventually force the decoder to select
a value that is valid but poorly aligned with the prompt.
To detect such situations, before assigning the current variable we ask the LLM whether
its feasible domain, as determined by the specification and the current partial
assignment, contains at least one value compatible with the prompt. A negative answer
indicates that, although the domain is formally feasible, the LLM considers all of its
values unsuitable for the prompt.
Its judgment is not an objective measure of user intent, but reflects the model’s own
assessment of the values that remain feasible.
When the LLM judges that none of the values in the current feasible domain adequately
satisfies the prompt, we say that a \emph{prompt conflict} (\pconflict) occurs. Although
the specification still admits one or more feasible values, the method backtracks and
revises earlier assignments responsible for restricting the current domain.
This allows the decoder to explore a different
specification-compliant trajectory instead of committing to a locally feasible
but undesirable value. 
Our focus is on resolving such mismatches once detected, rather than on
developing a general characterization of prompt adherence. In principle, the
relevant signal is the discrepancy between the constrained and unconstrained
distributions, since the latter reflects what the model would prefer to
generate in the absence of specification-based filtering. In particular, we
would like to avoid forcing the decoder to select tokens that have very low
probability under the unconstrained distribution.
In practice, we approximate this discrepancy using the domain probe described above, which
is simpler to integrate and proved effective in our experiments.

Consider again the laptop example, with a prompt requiring
$battery$ to be $standard$. The decoder may first assign
$CPU=i7$, although this is not required by the prompt. Under the constraints, this choice
restricts $RAM$ to $ram16$ or $ram32$, which in turn forces
$BATTERY=extended$. The resulting configuration satisfies the
formal specification but conflicts with the request in the prompt. Backtracking makes it
possible to revise the earlier $CPU$ assignment and search for an
alternative satisfying configuration.

When a \pconflict occurs, the current feasible domain is excluded from the specification,
causing the algorithm to reconsider earlier assignments.
For a variable $x$ and a domain $dom(x)$, let
$\phi_x(dom(x))$ denote a formula such that
$
    \mu \models \phi_x(dom(x))
    \quad\text{iff}\quad
    \mu(x) \in dom(x).
$
Conjoining $\phi_x(dom(x))$ with the specification restricts its satisfying
assignments to those in which $x$ takes a value in $dom(x)$. Consequently,
conjoining $\neg\phi_x(dom(x))$ excludes all satisfying assignments in which
$x$ takes a value in $dom(x)$.

Algorithm~\ref{alg:suregen-bt} extends
Algorithm~\ref{alg:suregenfw} by allowing the trail, the stack of unassigned
variables, and the generated output to be revised. At each step, the reasoner
computes the feasible domain
$dom_{\mathcal{F}_{k-1}}(x_k)$ of the current variable under the
partial assignment. 
The function $dom\_probe\_accepts(p, dom_{\mathcal{F}_{k-1}}(x_k))$
then returns true if the LLM judges that this domain contains at least one value compatible with the
prompt.
If so, constrained decoding selects a feasible
value and generation proceeds as in the forward-only method.
If the domain is rejected, the algorithm conjoins
$
\neg\phi_{x_k}\!\left(
dom_{\mathcal{F}_{k-1}}(x_k)
\right)
$
to $\mathcal{F}$. 
The updated specification, together with the current partial assignment, becomes
inconsistent, so assignments are retracted until satisfiability is restored and generation
can resume. If backtracking reaches the root level, the algorithm returns \punsat,
indicating that the accumulated prompt-based exclusions are jointly inconsistent with the
original specification.
This form of backtracking revises locally attractive assignments whose downstream
consequences conflict with the prompt. Unlike clause learning in
SMT~\cite{NieuwenhuisOT06}, which typically blocks specific combinations of assignments,
our blocking is global: once the current feasible domain of a variable is rejected, its
values are excluded from all subsequent branches. This avoids enumerating partial
assignments and supports real-valued or unbounded domains, where blocking assignments
individually may require infinitely many exclusions and cannot guarantee termination. For
finite domains, the number of value exclusions is bounded by the sum of the domain sizes,
whereas context-sensitive blocking may require exponentially many combinations. This is
particularly undesirable in our setting because exploring those combinations entails
additional LLM calls, which dominate the computational cost and are precisely what the
method aims to minimize.
This design deliberately trades conditioning prompt-based exclusions on earlier
assignments for bounded progress and is best suited to single-variable preferences,
similarly to interactive product configuration, where users choose a value for one
parameter at a time from the currently feasible options.

As with Algorithm~\ref{alg:suregenfw}, every assignment returned by
Algorithm~\ref{alg:suregen-bt} satisfies the formal specification. The algorithm
may additionally report \punsat{} when the accumulated prompt-based exclusions
eliminate every value of some variable.

\section{Experimental Evaluation}
\label{sec:exp}
In this section, we present an experimental evaluation of our approach, implemented in the
\suregen tool. \suregen supports a forward-only variant, \suregenfw (Algorithm~
\ref{alg:suregenfw}), and a backtracking variant, \suregenbt
(Algorithm~\ref{alg:suregen-bt}).
We evaluate whether \suregenfw reduces the cost of trial-and-error generation without
substantial runtime overhead, and whether \suregen-bt better preserves the LLM's own
interpretation of the prompt by resolving detected prompt conflicts through backtracking.

\paragraph{Benchmark}
We evaluate on 89 instances from four benchmark sources: 23
constraint-based configuration instances from CLib~\cite{subbarayan2004comparing}, 24
feature models from the S.P.L.O.T. collection~\cite{MendoncaBC09}, 9
configuration instances from COOM Suite~\cite{baumeister_2026}, and 33 instances from the QF\_LIA
and QF\_LRA logics of SMT-LIB~\cite{barrett2016smtlib}. 
We selected
instances that supported repeated feasible-domain analysis within our
experimental budget and whose variables admitted meaningful
natural-language interpretations for LLM-guided generation.
The
complete benchmark set covers product and software configuration, scheduling, routing, package
management, graph and combinatorial problems, and other numeric constraint tasks.
They include Boolean, enumerated,
integer, and real-valued variables, range from a few to several thousand
variables, and contain both propositional and arithmetic constraints. All
instances are represented in SMT-LIB~2 format~\cite{barrett2010smt} and solved
using Z3~\cite{de2008z3}.

\paragraph{Experimental setup}
All experiments were conducted using \texttt{gpt-oss-120b},
deployed with \texttt{vLLM} v0.23.0 on a single NVIDIA B300 SXM6 GPU with tensor parallelism set to
1. The model was served in its native MXFP4 format, with BF16 used for the remaining weights,
activations, and computation. The deployment was allocated 10 CPU cores and 60GiB of system memory.
Constrained decoding used \texttt{vLLM}'s automatic backend selection, with XGrammar as the primary
backend and LLGuidance as fallback. The maximum context length was 131072 tokens.

\paragraph{Compared methods and implementation details}

We implement \suregenfw and \suregenbt by assigning variables sequentially, using one
OpenAI-compatible API call per non-singleton domain; throughout the evaluation, we refer
to each such API request as an LLM call.
Each call receives a JSON Schema describing the current feasible domain and a prompt
containing the user requirements together with the previous assignments. The schema
supports Boolean, numeric, and enumerated values, including ranges and finite value sets,
consistently with existing constrained-decoding tools~\cite{lundberg2023guidance,
llamacpp2023}.
For Boolean and enumerated variables, feasible domains are computed directly
from Definition~\ref{def:feas-domain}; for integer and real variables, possibly
with unbounded domains, we use the procedure of~\cite{PrevitiIJM17}.
We compare against sequential rejection sampling (\srs), which mirrors token-level semantic
constrained decoding~\cite{Nagy2026ChopChop} at the variable level: proposed values are retained
only if the resulting partial assignment remains extendable to a satisfying assignment; otherwise,
they are blocked and resampled. 
We also consider \srnds, which
uses the same procedure but selects values uniformly at random, to assess the
benefit of LLM-guided value selection.
We additionally implemented complete rejection sampling (\rs), which generates one complete
assignment at a time, using structured output only, checks it against the formal specification, and
retries whenever the candidate is invalid. We tested \rs with temperatures $0.2$ and $0.5$ to
introduce sufficient variability and avoid repeatedly producing the same assignment. We also
considered variants of \rs and \srs in which the formal constraints are included in the prompt (\rsc
and \srsc). Since \rs was not competitive in runtime or number of solved instances, and adding the
constraints to the prompt provided no consistent reduction in the number of LLM calls while also
increasing runtime, we omit these variants from the main results. A detailed account of these
negative results is omitted for space.

\begin{figure}[t]
    \centering
    \includegraphics[width=0.98\columnwidth]{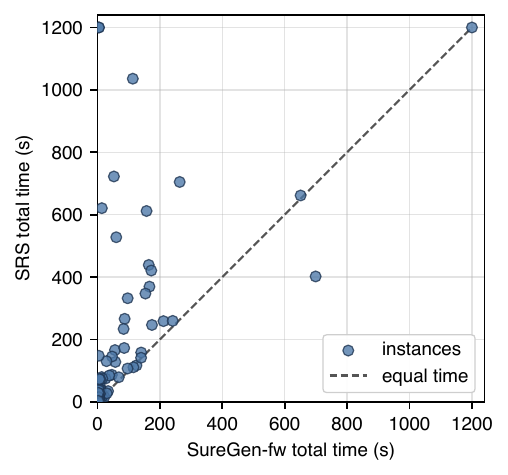}
    \caption{Runtime comparison between \suregenfw and \srs.} 
    \label{fig:runtime-scatter}
\end{figure} 

\begin{figure}[t]
    \centering
    \includegraphics[width=0.98\columnwidth]{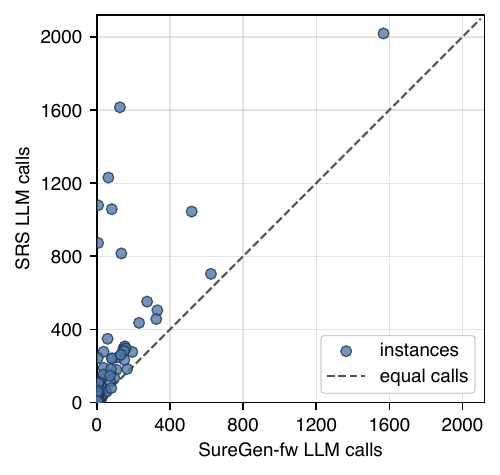}
    \caption{Comparison of the number of LLM calls made by \suregenfw and \srs}
    \label{fig:llm_calls_scatter}
\end{figure}

\subsection{Evaluation of Forward-Only Decoding}

This section evaluates whether specification-guided generation benefits from
shifting consistency reasoning to the symbolic solver rather than relying on
trial-and-error search. The latter may require many costly LLM calls and depends
on the model proposing effective values, which cannot be assumed in general.

We compare \suregenfw (Algorithm~\ref{alg:suregenfw}) with \srs using the same
fixed variable ordering and temperature $0$. 
For finite domains, both \suregenfw and \srs eventually find a solution whenever one exists.
However, \srs relies on value enumeration and therefore does not retain this guarantee for
real-valued or unbounded domains. In contrast, \suregenfw employs the procedure
of~\cite{PrevitiIJM17} to compute feasible domains without explicitly enumerating their values,
allowing it to handle real-valued and unbounded domains.
Figure~\ref{fig:runtime-scatter} compares the running times of \suregenfw and \srs using a timeout
of 1200 seconds per instance. \suregenfw is faster on most matched instances, often substantially
so.
\srs timed out on three instances, whereas \suregenfw timed out on one of them. Moreover, \srs
failed to return grammar-compliant values on four instances involving real-valued variables, which
may be due to the highly fragmented feasible domains produced by repeatedly blocking individual
values, making them harder for constrained-decoding tools to represent and enforce reliably.
Figure~\ref{fig:llm_calls_scatter} shows that \srs requires substantially more
LLM calls than \suregenfw, which helps explain the runtime difference.
In \srs, additional calls are needed whenever a proposed value is rejected by the solver.
By filtering the entire feasible domain before decoding, \suregenfw avoids these retries
and directly propagates singleton domains without invoking the LLM. Implied assignments
occurred in $96.6\%$ of successful runs, and the median instance had $51.0\%$ of its
variables assigned by implication.
The additional symbolic reasoning performed by \suregenfw remained modest: its
median share of total runtime was $1.1\%$, and it remained below $10\%$ on
$90.2\%$ of matched solved instances. Thus, shifting consistency reasoning from
repeated LLM interaction to the symbolic reasoner reduced inference calls
without introducing substantial overhead on the evaluated benchmarks.
These results also suggest that, once feasible-domain computation itself becomes
prohibitive, repeatedly invoking the LLM to explore and reject infeasible choices is
unlikely to improve overall scalability.

Finally, we compare \srs with \srnds to assess whether the LLM provides a useful value-selection
heuristic. While \srnds solved instances with small finite domains, it required more attempts and
failed on all QF\_LIA and QF\_LRA instances. These results suggest that the LLM can provide useful
guidance on the large or unbounded domains considered here, although this observation should not be
interpreted as a general guarantee. At the same time, this improved guidance does not translate into
better runtime performance, since LLM calls remain comparatively expensive.

\subsection{Evaluation of Backtracking}

The goal of backtracking is to improve consistency with the model’s judgment by revising
earlier assignments when they leave only values it considers unsuitable.
To evaluate this behavior, we focus on \pconflict{} events, which occur when the
domain probe judges that the current feasible domain contains no value compatible
with the user request and the current partial assignment.
In forward mode, a \pconflict{} is ignored, forcing the decoder to select a
value that the LLM itself judged unsuitable. Backtracking instead revises
earlier choices, aiming to produce a satisfying assignment that better reflects
the model's own interpretation of the prompt.
We compare the final assignments produced by the two modes using a post-hoc
model-based evaluator. The results therefore measure consistency with the
model's own stated preferences while preserving satisfaction of the formal
specification, rather than objective semantic optimality.
Because the domain probe is only an approximate preference signal, an incorrect
negative judgment may exclude a trajectory that would otherwise produce a valid
and intent-adherent assignment. Accordingly, \punsat{} denotes unsatisfiability
with respect to the accumulated probe-based exclusions, rather than definitive
unsatisfiability of the original prompt and specification.
For this evaluation, we instrument forward mode with the same domain probe used
by backtracking. Before each assignment, the probe evaluates the current
feasible domain under the partial assignment. Forward mode records and ignores
negative judgments, whereas backtracking attempts to resolve them by revising
previous choices.

\paragraph{Instance Selection}
We selected 35 of the 89 instances used in the preceding experiment. Whereas the previous experiment
evaluated the performance of our approach on instances containing as many as 3,189 variables, this
evaluation focuses on the quality of the generated results. We therefore restrict the analysis to
instances with at most 66 variables. In addition, a thorough quality assessment required running
each instance multiple times with different prompts, substantially increasing the total number of
experimental runs.

\paragraph{Results}
We evaluate output quality through two experiments. First, we construct prompts whose ground-truth
outcome is unsatisfiable and measure how often backtracking mode correctly returns \punsat. Second,
for each instance, we generate multiple prompts, obtain satisfying assignments from both forward and
backtracking modes, and use an LLM judge to determine whether each assignment conflicts with the
prompt.

\paragraph{\punsat test}
To construct ground-truth unsatisfiable prompts, we first computed a complete falsifying assignment
and then extracted a minimal falsifying assignment (MFA), defined as a falsifying assignment
involving a minimal number of variables. We translated each MFA into a natural-language prompt and
asked the backtracking algorithm to find a satisfying assignment.
Across 35 instances, backtracking mode correctly returned \punsat for 33. In the remaining two
cases, the domain probe failed to recognize that the feasible domains were incompatible with the
prompt and instead returned a solution. 
As expected, forward mode always returned a satisfying solution that, by construction,
conflicted with the prompt.

\paragraph{SAT test.}
In this experiment, we evaluate whether backtracking produces satisfying assignments that
better reflect the model’s interpretation of the user request.
We generated multiple prompts for each instance and retained
those for which both backtracking mode and forward mode encountered at least one intent conflict.
This yielded 98 instance--prompt pairs, covering all 35 original instances, with most instances
contributing more than one prompt.
To identify conflicts ignored by forward mode, we instrumented it with the same domain probe used by
backtracking mode. Before each assignment, the probe checked whether the current domain contained at
least one value compatible with the prompt. Unlike backtracking mode, however, forward mode ignored
negative judgments and still selected one of the available values.
We then collected the satisfying assignments produced by both modes and evaluated them with two LLM
judges, \texttt{gpt-oss-120b} and \texttt{gpt-5.5}. Each judge processed one variable at a time and
determined whether its assigned value conflicted with or adhered to the prompt. This post-hoc
evaluation is simpler than domain probing during generation, since it assesses a single assigned
value rather than an entire domain.
All assignments were generated using \texttt{gpt-oss-120b}, which was also used for domain probing.
Consequently, the judgments produced by \texttt{gpt-oss-120b} are not independent of the generation
process and may be affected by self-consistency bias. This evaluation is nevertheless consistent
with our goal of measuring adherence to the generating model's own interpretation of 
the prompt.
To test whether the observed effect also holds under a different evaluator, we additionally
judged the same assignments with \texttt{gpt-5.5}. We therefore interpret the \texttt{gpt-oss-120b}
results as measuring alignment with the generating model's stated preferences, and the
\texttt{gpt-5.5} results as complementary evidence that the improvement is not solely an artifact of
same-model evaluation.
Using \texttt{gpt-oss-120b} as the judge, 13 of the 98 assignments produced in backtracking mode
were classified as non-adherent to the prompt, compared with 92 of the 98 assignments produced in
forward mode. Using \texttt{gpt-5.5} as the judge, the corresponding counts were 14 and 90,
respectively. 
Thus, both judges found that backtracking substantially reduced assignments containing
values judged unsuitable for the prompt.

\section{Related Work}

While most existing approaches that combine large language models with reasoning focus on
translating natural-language requirements into a formal representation~\cite{YeCDD23,
WuICLR24, pmlr-v267-zhang25ds, PanAWW23}, our setting assumes that the formal problem is
already given and therefore targets a different, complementary problem.
For simplicity, we present the forward method for outputs consisting entirely of
assignments to variables in the specification. However, since its target is the
token-level specification-guided decoding problem of Definition~\ref{def:spec-decoding},
it can also constrain variables embedded in broader generation processes containing
variables outside the specification or content that cannot be fully formalized and may
influence the preferred constrained values. Methods that formalize and solve the complete
task therefore address a different setting.
Related to our approach, but addressing a different problem setting,
ChopChop~\cite{Nagy2026ChopChop} performs semantic constrained decoding for code generation by
checking the realizability of partial programs. As discussed in the experimental evaluation, its
token-level propose-and-reject procedure corresponds closely to \srs at the assignment level.
~\citet{BonlarronRMR25} combine symbolic reasoning with large language models for
constrained text generation. Unlike our setting, their constraints govern textual properties and are
introduced dynamically during generation, whereas ours encode semantic relationships over a fixed,
pre-specified set of variables.

\section{Conclusions and Future Work}

We introduced a specification-guided decoding approach that integrates symbolic
reasoning into LLM generation.
By incrementally refining the decoding grammar according to feasible values derived from a
logical specification, the method goes beyond syntactic validity by enforcing logical
dependencies among generated values and guaranteeing that every completed output satisfies
the specification by construction.
The approach is model-agnostic, requires no retraining, and builds directly on existing
grammar-constrained decoding techniques.

Experiments across diverse configuration and constraint-solving benchmarks show that
incorporating reasoning into generation reduces repeated LLM inference while
guaranteeing compliance with the formal specification.

We also introduced a backtracking extension that prevents forward decoding from selecting
values the LLM itself judges unsuitable merely because they remain feasible.

Both approaches depend on the tractability of the underlying reasoning tasks, and some
instances may exceed practical resource limits. However, formal specifications of
practical interest are not necessarily hard, and our experiments show that many can be
handled effectively. By filtering the entire feasible domain before invoking the LLM, our
design shifts most of the computational burden to the symbolic solver. This trade-off is
favorable because LLM inference is the main performance bottleneck, whereas solver calls
are comparatively inexpensive on the instances considered. Moreover, when complete domain
filtering is already intractable, methods requiring additional LLM calls or repeated
exploration of infeasible choices are unlikely to scale better.

Future work includes developing context-sensitive blocking strategies that retain
termination guarantees for real-valued or unbounded domains, and using token probabilities
or value rankings to improve domain evaluation.

\section*{Acknowledgments}

This work was supported by Vinnova, Sweden's innovation agency, the Swedish
Research Council, with support through the SAFIR project (Secure AI for
Intelligent Resilience and Confidentiality in the Cloud; reference
no.~2025-03039).

\bibliography{refs}

\appendix

\end{document}